\documentclass[12pt]{article}
\usepackage{framed,epsf,latexsym,amsmath,amssymb,amscd,textcomp}
\usepackage[numbers]{natbib}                                 
\usepackage{graphicx}
\usepackage[pdftex,colorlinks]{hyperref}
\usepackage[margin=1.0in]{geometry}

\usepackage{algorithm}
\usepackage[noend]{algorithmic}

\newtheorem{theorem}{Theorem}[section]
\newtheorem{lemma}[theorem]{Lemma}

\newtheorem{counter-example}[theorem]{Counter example}

\newtheorem{open question}[theorem]{Open question}
\newtheorem{corollary}[theorem]{Corollary}

\newtheorem{definition}[theorem]{Definition}

\newtheorem{claim}{Claim}

\newcommand{\ignore}[1]{}

\newcommand{\ca}{{\cal A}}
\newcommand{\cb}{{\cal B}}
\newcommand{\cd}{{\cal D}}

\newcommand{\ct}{{\cal T}}

\newcommand{\ch}{{\cal H}}

\newcommand{\cx}{{\cal X}}
\newcommand{\cy}{{\cal Y}}

\DeclareMathOperator*{\maj}{MAJ}

\newcommand{\DNF}{\mathrm{DNF}}

\newcommand{\x}{\mathbf{x}}
\newcommand{\z}{\mathbf{z}}

\newcommand{\proof}{{\par\noindent {\bf Proof}\space\space}}
\newcommand{\proofbox}{\hfill $\Box$}

\DeclareMathOperator{\poly}{poly}
\DeclareMathOperator*{\E}{\mathbb{E}}

\title{Distribution Free Learning with Local Queries\footnote{This paper is based on the M.Sc. thesis~\cite{bary2015learning} of the first author. The thesis offers a more elaborated discussion, as well as experiments.}}

\author{
Galit Bary-Weisberg\thanks{Matific inc. Most work was done while the author was an M.Sc. student at the Hebrew University, Jerusalem, Israel} \hspace{1cm}
Amit Daniely\thanks{Google inc. Most work was done while the author was a Ph.D. student at the Hebrew University, Jerusalem, Israel} \hspace{1cm}
Shai Shalev-Shwartz\thanks{School of Computer Science and Engineering, The Hebrew University, Jerusalem, Israel} 
}

\begin{document}
\maketitle
\setcounter{page}{0}

\thispagestyle{empty}
\maketitle

\begin{abstract}
The model of learning with \emph{local membership queries}
interpolates between the PAC model and the membership queries model by
allowing the learner to query the label of any example that is similar
to an example in the training set. This model, recently
proposed and studied by \citet{awasthi2012learning},  aims to facilitate
practical use of membership queries.

We continue this line of work, proving both positive and negative
results in the {\em distribution free} setting. We restrict to the
boolean cube $\{-1, 1\}^n$, and say that a query is $q$-local if it is of a hamming
distance $\le q$ from some training example. On the positive side, we
show that $1$-local queries already give an additional strength, and
allow to learn a certain type of DNF formulas. On the negative side,
we show that even $\left(n^{0.99}\right)$-local queries cannot help to
learn various classes including Automata, DNFs and more. Likewise,
$q$-local queries for any constant $q$ cannot help to learn Juntas, Decision Trees, Sparse
Polynomials and more. Moreover, for these classes, an algorithm that uses
$\left(\log^{0.99}(n)\right)$-local queries would lead to a
breakthrough in the best known running times.
\end{abstract}

\newpage

\section{Introduction} \label{intro}

% Intuitive introduction
A child typically learns to recognize a cat based on two types of input. The first is given by her parents, pointing at a cat and saying ``Look, a cat!". The second is given as a response to the child's frequent question ``What is that?".
% History
These two types of input were the basis for the learning model
originally suggested by \citet{valiant1984theory}.  Indeed, in
Valiant's model, the learner can randomly sample labelled
examples from ``nature", but it can also make a {\em membership query
  (MQ)} for the label of {\em any} unseen example. Today, the acronym
PAC stands for the restricted model in which MQ are forbidden, while
the full model is called PAC+MQ.  Much work has been done
investigating the limits and the strengths of MQ. In particular,
membership queries were proven stronger than the vanilla PAC
model~\citep{angluin1987learning, blum1992fast,
  bshouty1995exact,jackson1994efficient}. Yet, MQ are rarely used in
practice.
%Why not used?
This is commonly attributed to the fact that MQ algorithms query very artificial examples, that are uninterpretable by humans (e.g., \cite{baum1992query}).

% Local MQ
\citet{awasthi2012learning} suggested a solution to the problem of unnatural examples. They considered a mid-way model that allows algorithms to make only \textit{local} queries, i.e., query examples that are close to examples from the sample set. 
Hopefully, examples which are similar to natural examples will appear natural to humans.
Awasti et al. considered the case where the instance space is $\{-1, 1\}^n$ and the distance between examples is the Hamming distance. They proved positive results on learning sparse polynomials with $O(\log(n))$-local queries under what they defined as locally smooth distributions\footnote{A distribution is locally $\alpha$-smooth for $\alpha\ge 1$ if its density function is $\log(\alpha)$-Lipschitz.}.
They also proposed an algorithm that learns DNF formulas under the uniform distribution in quasi-polynomial time using $O(\log(n))$-local queries.

Our work follows Awasthi et al. and investigates local queries in the {\em distribution free} setting, in which no explicit assumptions are made on the underlying distribution, but only on the learned hypothesis. We prove both positive and negative results in this context:
\begin{itemize}
\item One of the strongest and most beautiful results in the MQ model shows that automata are learnable with membership queries~\cite{angluin1987learning}. We show that unfortunately, this is probably not the case with local queries. Concretely, we show that even $\left(n^{0.99}\right)$-local queries cannot help to learn automata. Namely, such an algorithm will imply a standard PAC algorithm for automata. As learning automata is hard under several assumptions~\cite{kearns1994cryptographic, daniely2014complexity}, our result suggests that it is hard to learn automata even with $\left(n^{0.99}\right)$-local queries.
\item We prove a similar result for several additional classes. Namely, we show that $\left(n^{0.99}\right)$-local queries cannot help to learn DNFs, intersection of halfspaces, decision lists, depth-$d$ circuits for any $d\ge 2$, and depth-$d$ threshold circuits for any $d\ge 2$. Likewise, for any constant $q$, $q$-local queries cannot help to learn Juntas, Decision Trees, Sparse polynomials, and Sparse polynomial threshold functions. In fact, we show that even $\left(\log^{0.99}(n)\right)$-local queries are unlikely to lead to polynomial time algorithms. Namely, any algorithm that uses $\left(\log^{0.99}(n)\right)$-local queries will result with a PAC algorithm whose running time significantly improves the state of the art in these well studied problems. 
\item On the positive side we show that already $1$-local queries are
  probably stronger than the vanilla PAC model. Concretely, we show
  that a certain simplistic but natural learning problem, which we term
  learning DNFs with evident examples, is learnable with $1$-local queries. Furthermore, we show that
  without queries, this problem is at least as hard as learning
  decision trees, that are conjectured to be hard to learn.
\end{itemize}

\section{Previous Work} \label{prev}

\paragraph{Membership Queries}
Several concept classes are known to be learnable only if membership queries are allowed:
Deterministic Finite Automatons \citep{angluin1987learning}, k-term DNF for $k= \frac{\log(n)}{\log(\log(n))}$ \citep{blum1992fast}, decision trees and k-almost monotone-DNF formulas \citep{bshouty1995exact}, intersections of k-halfspaces~\citep{baum1991neural} and DNF formulas under the uniform distribution \citep{jackson1994efficient}. The last result builds
on Freund's boosting algorithm \citep{freund1995boosting} and the Fourier-based technique for learning using membership queries due to \citep{kushilevitz1993learning}.
We note that there are cases in which MQ do not help. E.g., in the case of learning DNF and CNF formulas \citep{angluin1995won}, assuming that one way functions exist, and in the case of distribution free agnostic learning \citep{feldman2009power}.

\paragraph{Local Membership Queries}
Awasthi et al. focused on learning with $O(\log(n))$-local queries. They showed that $t$-sparse polynomials are learnable under locally smooth distributions using $O\left(\log(n)+\log(t) \right)$-local queries, and that DNF formulas are learnable under the uniform distribution in quasi-polynomial time ($n^{O(\log\log n})$) using $O(\log(n))$-local queries. They also presented some results regarding the strength of local MQ. They proved that under standard cryptographic assumptions,  $(r+1)$-local queries are more powerful than $r$-local queries (for every $1\leq r\leq n-1$). They also showed that local queries do not always help. They showed that if a concept class is agnostically learnable under the uniform distribution using $k$-local queries (for constant $k$) then it is also agnostically learnable (under the uniform distribution) in the PAC model.

We remark that besides local queries, there were additional suggestions to solve the problem of unnatural examples. For example, to allow ``I don't know" answers, or to be tolerant to some incorrect answers ~\cite{angluin1994randomly, angluin1997malicious,blum1995learning,sloan1994learning,bisht2008learning}.

\section{Setting} 
We consider binary classification where the instance space is $\cx=\cx_n=\{-1, 1\}^n$ and the label space is $\cy=\{0,1\}$.
A learning problem is defined by a hypothesis class $\ch\subset \{0,1\}^{\cx}$. 
The learner receives a \emph{training set}
\[
S = \{(\x_1,h^\star(\x_1)) , (\x_2,h^\star(\x_2)) , \ldots , (\x_m,h^\star(\x_m)) \} \in (\cx \times \cy)^m
\]
where the $\x_i$'s are sampled i.i.d. from some {\em unknown} distribution $\cd$ on $\cx$ and $h^\star:\cx\to\cy$ is some {\em unknown} hypothesis. 
The learner is also allowed to make membership queries. Namely, to call an oracle, which receives as input some $\x \in \cx$ and returns $h^\star(\x)$. We say that a membership query for $\x\in\cx$ is {\em $q$-local} if there exists a training example $\x_i$ whose Hamming distance from $\x$ is at most $q$. 

The learner returns (a description of) a hypothesis $\hat{h} : \cx \rightarrow \cy$.
The goal is to approximate $h^\star$, namely to find $\hat{h}:\cx\to \cy$ with {\em loss} as small as possible, where the loss is defined as $L_{\cd,h^\star}(\hat{h})=\Pr_{\x\sim\cd}\left(\hat{h}(\x)\ne h^\star(\x)\right)$. 
We will focus on the so-called realizable case where $h^\star$ is assumed to be in $\ch$, and will require algorithms to return a hypothesis with loss $< \epsilon$ in time that is polynomial in $n$ and $\frac{1}{\epsilon}$. Concretely,

\begin{definition} [Membership-Query Learning Algorithm] \label{MQ-alg}
We say that a learning \\ algorithm $\ca$ {\em learns $\ch$ with $q$-local membership queries} if
\begin{itemize}
\item
There exists a function $m_\ca \left(n,\epsilon\right)\le \poly\left(n,\frac{1}{\epsilon}\right)$, such that 
for every distribution $\cd$ over $\cx$, every $h^{\star}\in\ch$ and every $\epsilon>0$, if $\ca$ is given access to $q$-local membership queries, and a training sequence
\[
S = \{(\x_1,h^\star(\x_1)) , (\x_2,h^\star(\x_2)) , \ldots , (\x_m,h^\star(\x_m)) \} 
\]
where the $\x_i$'s are sampled i.i.d. from $\cd$ and $m \ge m_\ca (n,\epsilon)$, then with probability of at least\footnote{The success probability can be amplified to $1-\delta$ by repetition.} $\frac{3}{4}$ (over the choice of $S$), the output $\hat{h}$ of $\ca$ satisfies 
$L_{\cd , h^\star}(\hat{h}) < \epsilon$.
\item Given a training set of size $m$
\begin{itemize}
\item $\ca$ asks at most $\poly(m,n)$ membership queries.
\item $\ca$ runs in time $\poly(m,n)$.
\item The hypothesis returned by $\ca$ can be evaluated in time $\poly(m,n)$.
\end{itemize}
\end{itemize}
\end{definition} 
We remark that learning with $0$-local queries is equivalent to PAC learning, while learning with $n$-local queries is equivalent to PAC+MQ learning.

\section{Learning DNFs with Evident Examples}
Intuitively, when evaluating a DNF formula on a given example, we check a few conditions corresponding to the formula's terms, and deem the example positive if one of them holds.
We will consider the case that for each of these conditions, there is a chance to see a ``prototype example", that satisfies it in a strong or evident way. In the sequel, we denote by $h_F : \{-1,1\}^n \to \{0,1\}$ the function induced by a DNF formula $F$ over $n$ variables.
\begin{definition} 
\label{satisfies evidently}
Let $F=T_1 \vee T_2 \vee \ldots \vee T_d$ be a DNF formula. We say that an example $\x\in \{-1,1\}^n$ satisfies a term $T_i$ (with respect to the formula $F$) {\bf evidently} and denote $T_i(\x)\equiv 1$ if :
\begin{itemize}
\item
It satisfies $T_i$. (In particular, $h_F(\x)=1$)
\item
It does {\bf not} satisfy any other term $T_k$ (for $k\neq i$) from F.
\item
No coordinate change will turn $T_i$ False and another term $T_k$ True.
Concretely, if for $j\in [n]$ we denote $\x^{\oplus j} = (x_1, \ldots, x_{j-1}, -x_j, x_{j+1}, \ldots ,x_n)$, then for every coordinate $j\in [n]$, if $\x^{\oplus j}$ satisfies $F$ (i.e. if $h_F(\x^{\oplus j} )=1$) then $\x^{\oplus j}$ satisfies $T_i$ and only $T_i$. 
\end{itemize}
\end{definition} 

\begin{definition} 
\label{distribution weakly realized by DNF with evident }
Let $F=T_1 \vee T_2 \vee \ldots \vee T_d$ be a DNF formula. We say that  $h^\star : \{-1,1\}^n \to \{0,1\}$ is {\em realized by $F$ with evident examples} w.r.t. a distribution $\cd$ if $h^\star = h_F$ and for every term\footnote{We note that the quantity $\frac{1}{n}$ in the following equation is arbitrary, and can be replaced by $\frac{1}{n^c}$ for any $c>0$.} $T_i$, $\Pr_{\x\sim\cd}\left(T_i(\x)\equiv 1|T_i(\x)=1\right)\ge\frac{1}{n}$.
\end{definition} 
For example, the definition holds whenever $h^\star$ is realized by a DNF in which any pair of different terms contains two opposite literals.
Also, functions that are realized by a decision tree can be also realized by a DNF in which every positive example satisfies a single term, corresponding to root-to-leaf path that ends with a leaf labelled by $1$.
Hence, the assumption holds for decision trees provided that for every such path, there is a chance to see an example satisfying it evidently, in the sense that flipping each variable in the path will turn the example negative.

\subsection{An algorithm}

\begin{algorithm}[ht]
\caption{}
\textbf{Input: } $S_1 , S_2 \in (\{-1,1\}^n  \times\{0,1\})^m $\\
\textbf{Output: } A $\DNF$ formula $H$
\begin{algorithmic}\label{alg2}

\STATE Start with an empty $\DNF$ formula $H$

\FORALL {positive examples $(\x,y) \in S_1$}
\STATE Define $T = x_1\wedge \overline{x_1}\wedge x_2\wedge \overline{x_2}\wedge \ldots
\wedge x_n 	\wedge \overline{x_n} $ \FOR{$ 1\leq j \leq n$}
\STATE Query $\x^{\oplus j}$ to get $h^\star(\x^{\oplus j})$
\IF{$h^\star(\x^{\oplus j})=1$}
\STATE Remove $x_j$ and $\overline{x_j}$ from $T$
\ENDIF
\IF{$h^\star(\x^{\oplus j})=0$}
\IF{$x_j=1$}
\STATE Remove $\overline{x_j}$ from $T$
\ENDIF
\IF{$x_j=0$}
\STATE Remove $x_j$ from $T$
\ENDIF
\ENDIF

\ENDFOR
\STATE $H = H \vee T$
\ENDFOR
	
\FORALL{$T$ in $H$}
\IF{$T(\x)=1$ but $y=0$ for some $(\x,y) \in S_2$}
\STATE Remove $T$ from $H$
\ENDIF
\ENDFOR
\STATE Return {H}

\end{algorithmic}
\end{algorithm}

\begin{theorem}
\label{learnable weak evident dist} 
Algorithm \ref{alg2} learns with $1$-local-queries poly-sized $\DNF$s with evident examples.
\end{theorem}
{\bf Idea} The algorithm is based on the following claim that follows easily from definition~\ref{satisfies evidently}.
\begin{claim}\label{propeq}
Let $F=T_1 \vee T_2 \vee \ldots \vee T_d$ be a $\DNF$ formula over $\{-1,1\}^n$. 
For every $\x \in\{-1,1\}^n$ that satisfies a term $T_i$ evidently (with respect to $F$), and for every $j \in [n]$ it holds that:
$$
h_F(\x^{\oplus j} )=1 \Longleftrightarrow  \text{ the term $T_i$ does not contain the variable } x_j
$$
\end{claim}
By this claim, if $\x$ is an evident example for a certain term $T$, one can easily reconstruct $T$. Indeed, by flipping the value of each variable and checking if the label changes, one can infer which variables appear in $T$. Furthermore, the sign of these variables can be inferred from their sign in $\x$. Hence, after seeing an evident example for all terms, one can have a list of terms containing all terms in the DNF. This list might have terms that are not part of the DNF. Yet, such terms can be thrown away later by testing if they make wrong predictions. 

\medskip

\proof (of Theorem \ref{learnable weak evident dist})
We will show that algorithm \ref{alg2} learns with $1$-local-queries any function that is realized by a $\DNF$ of size $\le n^2$ with evident examples. Adapting the proof to hold with $n^c$ instead of $n^2$, for any $c>0$, is straight-forward.
First, it is easy to see that this algorithm is efficient. Now, fix a distribution $\cd$ and let $h^\star :  \{-1,1\}^n \to  \{0,1\}$ be a hypothesis that is realized, w.r.t. $\cd$, by a $\DNF$ formula $F=T_1 \vee T_2 \vee \ldots \vee T_d,\;d\le n^2$ with evident examples. Let $\epsilon > 0$, and suppose we run the algorithm on two indepent samples from $\cd$, denoted $S_1 = \{(\x_i,h^\star(\x_i)\}_{i=1}^{m_1}$ and $S_2 = \{(\x'_i,h^\star(\x'_i)\}_{i=1}^{m_2}$. We will show that if  $m_1 \ge \frac{32n^3}{\epsilon} \log \frac{32n^2}{\epsilon} \geq \frac{32nd}{\epsilon} \log \frac{32d}{\epsilon}$ and $m_2\geq \frac{32m_1}{\epsilon} \log \frac{32m_1}{\epsilon}$ then with probability of at least $\frac{3}{4}$, the algorithm will return a function $\hat h$ with $L_{\cd,h^\star}(\hat{h}) < \epsilon$.
Let $H$ be the $\DNF$ formula returned by the algorithm, and let $\hat{h}$ be the function induced by $H$. We have that
\begin{eqnarray*}
L_{\cd,h^\star}(\hat{h}) &=& \Pr_{\x\sim\cd}\left(h^\star(\x)\ne \hat{h}(\x)\right)
\\
&=& \Pr_{\x\sim\cd}\left(h^\star(\x) = 1\text{ and } \hat{h}(\x)=0\right) + 
\Pr_{\x\sim\cd}\left(h^\star(\x) = 0\text{ and } \hat{h}(\x)=1\right)
\end{eqnarray*}
The proof will now follow from claims \ref{claim_1} and \ref{claim_2}.
\begin{claim}\label{claim_1}
With probability at least $\frac{7}{8}$ over the choice of $S_1,S_2$ we have
\begin{equation}\label{eq:1}
\Pr_{\x\sim\cd}\left(h^\star(\x) = 1\text{ and } \hat{h}(\x)=0\right)\le \frac{\epsilon}{2}
\end{equation}
\end{claim}
\proof 
We first note that if there is an evident example $\x$ for a term $T_i$ in $S_1$, then $T_i$ will be in the output formula. Indeed, in the for-loop that go over the examples in $S_1$, when processing the example $(\x,h^\star(\x))$, it is not hard to see that $T_i$ will be added. We furthermore claim that the term $T_i$ won't be removed at the for loop that tests the terms collected in the first loop. Indeed, if for some $(\x,h^\star(\x))\in S_2$ we have $T_i(\x)=1$, it must be the case that $h^\star(\x)=1$.
Now, we say that the term $T_i$ is {\em revealed} if we see an evident example for this term in $S_1$. We also denote $p_i = \Pr_{\x\sim\cd}\left(T_i(\x)=1\right)$. We have
\begin{eqnarray*}
\Pr_{\x\sim\cd}\left(h^\star(\x) = 1\text{ and } \hat{h}(\x)=0\right) &\le& \sum_{i=1}^d \Pr_{\x\sim\cd}\left(T_i(\x) = 1\text{ and } \hat{h}(\x)=0\right)
\\
&\le& \sum_{i:T_i\text{ is not revealed}} p_i
\end{eqnarray*}
Now, by the assumption that $h^\star$ is realized with evident queries, the probability (over the choice of $S_1,S_2$) that $T_i$ is not revealed is at most $\left(1-\frac{p_i}{n}\right)^{m_1}$. Hence, if we denote by $A_i$ the event that $T_i$ is not revealed, we have
\begin{eqnarray*}
\E_{S_1 \sim \cd^{m_1}}\left[\Pr_{\x\sim\cd}\left(h^\star(\x) = 1\text{ and } \hat{h}			
(\x)=0\right)\right] &\leq& {\E}_{S_1 \sim \cd^{m_1}}\left[\sum_{i=1}^d p_i \cdot {1}_{A_i}\right]
\\
&=& \sum_{i=1}^d p_i {\E}_{S_1 \sim \cd^{m_1}}[{1}_{A_i}]
\\
&=& \sum_{i=1}^d p_i {\Pr}_{S_1 \sim \cd^{m_1}}[A_i]
\\
&=& \sum_{i=1}^d p_i \left(1-\frac{p_i}{n}\right)^{m_1}
\\
&=&\sum_{i | p_i<\frac{\epsilon}{32d}} p_i \left(1-\frac{p_i}{n}\right)^{m_1} + \sum_{i | 
p_i\geq\frac{\epsilon}{32d}} p_i  \left(1-\frac{p_i}{n}\right)^{m_1}
\\
&\leq & \sum_{i | p_i<\frac{\epsilon}{32d}} \frac{\epsilon}{32d} + \sum_{i | 
p_i\geq\frac{\epsilon}{32d}}  \left(1-\frac{p_i}{n}\right)^{m_1}
\\
&\leq & d \cdot \frac{\epsilon}{32d} + \sum_{i | p_i\geq\frac{\epsilon}{32d}} e^{-\frac{m_1p_i}
{n}}
\\
&\leq&  \frac{\epsilon}{32} + d \cdot e^{-\frac{m_1\epsilon}{32dn}} 
\end{eqnarray*}		
Since $m_1 \geq \frac{32dn}{\epsilon} \log \frac{32d}{\epsilon}$  the last expression is bounded by $\frac{\epsilon}{16}$. By Markov's inequality we conclude that the probabilty over the coice of $S_1,S_2$ that (\ref{eq:1}) does not lold is less than $\frac{1}{8}$.
\proofbox

\begin{claim}\label{claim_2}
With probability at least $\frac{7}{8}$ over the choice of $S_1,S_2$ we have
\begin{equation}\label{eq:2}
\Pr_{\x\sim\cd}\left(h^\star(\x) = 0\text{ and } \hat{h}(\x)=1\right)\le \frac{\epsilon}{2}
\end{equation}
\end{claim}
\proof
Let $\hat{T}_1,\ldots,\hat{T}_{r}$ be the terms that were added to $H$ at the first for-loop. Denote $q_i = \Pr_{\x\sim\cd}\left(\hat{T}_i(\x) = 1\text{ and } h^{\star}(\x)=0\right)$.
We have
\begin{eqnarray*}
\Pr_{\x\sim\cd}\left(h^\star(\x) = 0\text{ and } \hat{h}(\x)=1\right) &\le& \sum_{i\;:\;\hat{T}_i\text{ is not removed}} q_i
\end{eqnarray*}
Now, the probability that $\hat{T}_i$ is not removed is $\left(1-q_i\right)^{m_2}$. Hence, using an argument similar to the one used in the proof of claim \ref{claim_1}, and since $m_2 \geq \frac{32m_1}{\epsilon} \log \frac{32m_1}{\epsilon} \geq \frac{32r}{\epsilon} \log \frac{32r}{\epsilon}$, the claim follows.
\proofbox

\proofbox

\subsection{A matching Lower Bound}

In this section we provide evidence that the use of queries in our algorithm is crucial. We will show that the problem of learning poly-sized decision trees can be reduced to the problem of learning DNFs with evident examples. As learning decision trees is conjectured to be intractable, this reduction serves as an indication that learning DNFs with strongly evident examples is hard without membership queries. In fact, we will show that learning decision trees can be even reduced to the case that {\em all} positive examples are evident.

\begin{theorem}\label{hardness}
Learning poly-sized DNFs with evident examples is as hard as learning poly-sized decision trees.
\end{theorem}
We denote by $h_T$ the function induced by a decision tree $T$. The proof will use the following claim:
\begin{claim} \label{reduction from DT to DNF}
There exists a mapping (a reduction) $\varphi : \{-1,1\}^n \rightarrow \{-1,1\}^{2n} $, that can be evaluated in $poly(n)$ time so that for every decision tree $\ct$ over $\{-1,1\}^n$ there exists a $\DNF$ formula $F$ over $\{-1,1\}^{2n}$ such that the following holds:
\begin{enumerate}
\item
The number of terms in $F$ is upper bounded by the number of leaves in $\ct$
\item
$h_\ct = h_F \circ \varphi$
\item
$\forall \x$ such that $h_\ct(\x) = 1$ ,  $\varphi (\x)$ satisfies some term in $F$ evidently. 
\end{enumerate}
\end{claim}

\proof
Define $\varphi$ as follows: 
$$ \forall \x=(x_1,x_2, \ldots , x_n) \in \cx_n  \qquad \varphi(x_1,x_2, \ldots , x_n) = (x_1,x_1,x_2, x_2 , \ldots , x_n, x_n) $$
Now, for every tree $\ct$, we will build the desired $\DNF$ formula $F$ as follows: 
First we build a $\DNF$ formula $F'$ over  $\{-1,1\}^n$ . Every leaf labeled '$1$' in $\ct$ will define the following term- take the path from the root to that leaf and form the logical AND of the literals describing the path. $F'$ will be a disjunction of these terms. Now, for every term $T$ in $F'$ we will define a term $\phi(T)$ over $\cx_{2n} $ in the following way: Let $P_T = \{i \in [n] : x_i \textit{ appear in T} \}$ and $N_T = \{i \in [n] : \overline{x_i} \textit{ appear in T} \}$. So 
$$ T = \bigwedge_{j \in P_T} x_j \bigwedge_{j \in N_T} \overline{x_j}$$ 
Define 
\begin{gather*}
\phi(T) = \bigwedge_{j \in P_T} x_{2j-1} \bigwedge_{j \in P_T} {x_{2j}}
\bigwedge_{j \in N_T}\overline{x_{2j-1}} \bigwedge_{j \in N_T}  \overline{x_{2j}} \\
 \end{gather*}
Finally, define $F$ to be the $\DNF$ formula over $\cx_{2n}$ given by
$$ F =  \bigvee_{T \in F'} \phi(T)$$
We will now prove that $\varphi$ and $F$ satisfy the required conditions.
First, $\varphi$ can be evaluated in linear time in $n$.
Second, it is easy to see that $h_\ct = h_F \circ \varphi$, and as every term in $F$ matches one of $\ct$'s leaves, the number of terms in $F$ cannot exceed the number of leaves in $\ct$. It is left to show that the third requirement holds. Let there be an $\x$ such that $h_\ct(\x) = 1$, then $\x$ is matched to one and only one path from $\ct$'s root to a leaf labeled '1'. From the construction of $F$, $\x$ satisfies one and only one term in $F'$.
Regarding the last requirement, that no coordinate change will make one term from $F$ False and another one True, we made sure this will not happen by ``doubling" each variable. By this construction, in order to change a term from False to True at least two coordinate must change their value.
\proofbox

\medskip
\noindent We are now ready to prove theorem \ref{hardness}.
\medskip

\proof [of theorem~\ref{hardness}] 
Suppose that $\ca$ learns size-$n$ DNFs with evident examples. Using the reduction from claim~\ref{reduction from DT to DNF} we will build an efficient algorithm $\cb$ that learns size-$n$ decision trees. For every training set
\[
S = \{(\x_1,h^\star(\x_1)) , (\x_2,h^\star(\x_2)) , \ldots , (\x_m,h^\star(\x_m)) \} \in (\cx_n \times \{0,1\})^m
\]
we define 
\[
\varphi(S) := \{(\varphi(\x_1),h^\star(\x_1)) , (\varphi(\x_2)),h^\star(\x_2)) , \ldots , (\varphi(\x_m),h^\star(\x_m)) \} \in (\cx_{2n} \times \{0,1\})^m
\]
The algorithm $\cb$ will work as follows: Given a training set $S$, $\cb$ will run $\ca$ on $\varphi(S)$, and will return $\hat{h} \circ \varphi$, where $\hat{h}$ is the hypothesis returned by $\ca$. 
Since $\varphi$ can be evaluated in $poly(n)$ time and $\ca$ is efficient, $\cb$ is also efficient.
We will prove that $\cb$ learns size-$n$ trees. Since $\ca$ learns size-$n$ DNFs with evident examples, there exists a function $m_\ca \left(n,\epsilon\right)\le \poly\left(n,\frac{1}{\epsilon}\right)$, such that if $\ca$ is given a training sequence
\[
S = \{(\x_1,h^\star(\x_1)) , (\x_2,h^\star(\x_2)) , \ldots , (\x_m,h^\star(\x_m)) \} \in (\cx_n \times \{0,1\})^m
\]
where the $\x_i$'s are sampled i.i.d. from a distribution $\cd$, $h^\star$ is realized by a poly-sized DNF with evident examples, and $m \ge m_\ca (n,\epsilon)$, then with probability of at least $\frac{3}{4}$ (over the choice of $S$), the output $\hat{h}$ of $\ca$ satisfies 
$L_{\cd, h^\star}(\hat{h}) \le \epsilon$. 
Let $\cd$ be a distribution on $\cx_n$ and let $h_\ct$ be a hypothesis that can be realized by a tree with $\le n$ leafs. Define a distribution $\tilde{\cd}$ on $\cx_{2n}$ by,
\[
\tilde{(\cd)}(\z) =
\begin{cases} 
\cd(\x) & \textrm{if}~ \exists \x \in \cx_{n} \textrm{ such that}~ \z = \varphi(\x) \\
0 & \textrm{otherwise}~ 
\end{cases}
\] 
Now, since $h_\ct$ is realized by $\ct$, from the conditions that $\varphi$ satisfies, we get that $h_\ct= h \circ \varphi$, where $h$ is realized by a DNF of size $\le n$ with evident examples w.r.t. $\tilde{\cd}$.
Now if $S \in (\cx_n \times \{0,1\})^m $ is an i.i.d. sample with $m \ge  m_\ca (2n,\epsilon)$ we have that with probability of at least $\frac{3}{4}$ it holds that

\begin{eqnarray*}
L_{\cd,h_\ct}(\cb(S)) &=& L_{\cd,h_\ct}(\hat{h} \circ \varphi)
\\
&=& \mathop{\Pr}_{\x \sim \cd}[h_\ct(\x) \neq \hat{h} \circ \varphi(\x)]
\\
&=& \mathop{\Pr}_{\x \sim \cd}[h \circ \varphi (\x) \neq \hat{h} \circ \varphi(\x)] 
\\
&=& \mathop{\Pr}_{\z \sim \tilde{\cd}}[h(\z) \neq \hat{h}(\z)]
\\
&=& L_{\tilde{\cd},h}(\hat{h})
\\ 
&=& L_{\tilde{\cd},h}(\ca(\varphi(S))) < \epsilon
\end{eqnarray*}

\proofbox

\section{Lower Bounds}
We first present a general technique to prove lower bounds on learning with local queries.
For $A\subset \cx_n$ and $q>0$ denote
\[
B(A,q) = \{\x\in\cx_n\mid \exists \mathbf{a}\in A,\;d(\x,\mathbf{a}) \le q\}
\]
We say that a mapping $\varphi:\{-1, 1\}^n\to \{-1, 1\}^{n'}$ is a {\em $q$-reduction of type A} from a class $\ch$ to a class $\ch'$ if the following holds:
\begin{enumerate}
\item $\varphi$ is efficiently computable.
\item For every $h\in\ch$ there is $h'\in\ch'$ such that
\begin{enumerate}
\item $h=h'\circ\varphi$
\item\label{res_req} The restriction of $h'$ to $B(\varphi(\cx_n),q)\setminus \varphi(\cx_n)$ is the constant function $1$.
\end{enumerate}
\end{enumerate}
We say that a mapping $\varphi:\{-1, 1\}^n\to \{-1, 1\}^{n'}$ is a {\em $q$-reduction of type B} if the same requirements hold, except that (\ref{res_req}) is replaced by the following condition: For every $\z\in B(\varphi(\cx_n),q)$ there is a unique $\x\in \cx_n$ satisfying $d(\z,\varphi(\x))\le q$, and furthermore, $h'(\z)=h
(\x)$.

\begin{lemma}\label{lem:basic_lower_bound}
Suppose that there is a $q$-reduction $\varphi$ from $\ch$ to $\ch'$. Then, learning $\ch'$ with $q$-local queries is as hard as PAC learning $\ch$.
\end{lemma}
\proof (sketch)
Suppose that $\ca'$ learns $\ch'$ with $q$-local queries. We need to show that there is an algorithm that learns $\ch$. Indeed, by an argument similar to the one in theorem \ref{hardness}, it is not hard to verify that the following algorithm learns $\ch$. Given a sample $\{(\x_1,y_1),\ldots,(\x_m,y_m)\}\subset \{-1, 1\}^n\times\{0,1\}$, run $\ca'$ on the sample $\{(\varphi(\x_1),y_1),\ldots,(\varphi(\x_m),y_m)\}\subset \{-1, 1\}^{n'}\times\{0,1\}$. Whenever $\ca'$ makes a $q$-local query for $\z\in\{-1, 1\}^{n'}$, respond $1$ if $\varphi$ is of type A. If $\varphi$ is of type B, respond $y_i$, where $\x_i$ is the unique training sample satisfying $d(\z,\varphi(\x_i))\le q$. Finally, if $\ca'$ returned the hypothesis $h'$, return $h'\circ\varphi$.
\proofbox

\noindent We next use Lemma \ref{lem:basic_lower_bound} to prove that for several classes, local queries are not useful. Namely, if the class is learnable with local queries then it is also learnable without queries. We will use the following terminology. We say that $q$-local queries cannot help to learn a class $\ch$ if $\ch$ is learnable if and only if it is learnable with $q$-local queries.

\begin{corollary}\label{cor:DNF}
For every $\epsilon_0>0$, $\left(n^{1-\epsilon_0}\right)$-local queries cannot help to learn poly-sized DNFs, intersection of halfspaces, decision lists, depth-$d$ circuits for any $d=d(n)\ge 2$, and depth-$d$ threshold circuits for any $d=d(n)\ge 2$.
\end{corollary}
\proof
We will only prove the corollary for DNFs. The proofs for the
remaining classes are similar. Also, for simplicity, we will assume
that $\epsilon_0=\frac{1}{3}$. Consider the mapping $\varphi:\{-1, 1\}^n\to \{-1, 1\}^{n^3}$ that replicates each coordinate $n^2$ times. To establish the corollary, we show that $\varphi$ is an $\left((n')^{\frac{2}{3}}-1\right)$-reduction of type A from poly-sized DNF to poly-sized DNF.

Indeed, let $F=T_1\vee\ldots\vee T_d$ be a DNF formula on $n$ variables, consider the following formula on the $n^3$ variables $\{x_{i,j}\}_{1\le i \le n,1\le j\le n^2}$. 
Let,
\begin{eqnarray*}
T'_t(\{x_{i,j}\}_{i,j}) &=& T_t(x_{1,1},\ldots,x_{n,1})
\\
G'(\{x_{i,j}\}_{i,j}) &=& \vee_{i=1}^n \vee_{j=1}^{n^2-1} \left(x_{i,j} \wedge \neg x_{i,j+1}\right)\vee \left(x_{i,j+1} \wedge \neg x_{i,j}\right)
\\
F' &=& \left(T'_1\vee\ldots\vee T'_d\right) \vee G' 
\end{eqnarray*}
It is not hard to verify that $h_F = h_{F'} \circ \varphi$. Moreover, if $\x=\{x_{i,j}\}_{i,j}\in B(\varphi(\cx_n),n^2-1)\setminus \varphi(\cx_n)$ then $x_{i,j}\ne x_{i,j+1}$ for some $i,j$, meaning that $h_G(x)=1$ and therefore also $h_F(x)=1$.
\proofbox

\begin{corollary}\label{cor:Automata}
For every $\epsilon_0>0$, $\left(n^{1-\epsilon_0}\right)$-local queries cannot help to learn poly-sized automata.
\end{corollary}
\proof
Again, for simplicity, we assume that $\epsilon_0=\frac{1}{3}$, and consider the mapping $\varphi:\{-1, 1\}^n\to \{-1, 1\}^{n^3}$ that replicates each coordinate $n^2$ times. To establish the corollary, we show that $\varphi$ is an $\left((n')^{\frac{2}{3}}-1\right)$-reduction of type A from poly-sized automata to poly-sized automata.

Indeed, let $A$ be an automaton on $n$ variables. 
It is enough to show that there is an automaton $A'$ on the  $n^3$ variables $\{x_{i,j}\}_{1\le i \le n,1\le j\le n^2}$ such that (i) the size of $A'$ is polynomial, (ii) $A'$ accepts any string in $B(\varphi(\cx_n),n^2-1)\setminus \varphi(\cx_n)$, and (iii) $A'$ accepts $\varphi(x)$ if and only if $A$ accepts $x$. Now, by the product construction of automatons~\cite{sipser2006introduction}, if $A'_1,A'_2$ are automata that induce the functions $h_{A'_1},h_{A'_2}:\{-1, 1\}^{n^3}\to \{0,1\}$, then the function $h_{A'_1}\vee h_{A'_2}$ can be induced by an automaton of size $|A'_1|\cdot |A'_2|$. Hence, it is enough to show that there are poly-sized automata $A'_1, A'_2$ that satisfies (ii) and (iii) respectively.

A construction of a size $O(n^2)$ automaton that satisfies (ii) is a simple exercise. We next 
explain how to construct a poly-sized automaton $A'_2$ satisfying (iii). The states of $A'_2$ 
will be the $S(A)\times [n^2]$ (here, $S(A)$ denotes the set of states of $A$). The start state of 
$A'_2$ will be $(\alpha_0,1)$, where $\alpha_0$ is the start state of $A$, and the accept 
states of $A'_2$ will be the cartesian product of the accept states of $A$ with $[n^2]$. 
Finally if $\delta: S(A)\times \{-1, 1\}\to S(A)$ is the transition function of $A$, then the 
transition function of $A'_2$ is defined by
\[
\delta'((\alpha, i),b)=\begin{cases}
(\alpha, i+1) & 1\le i < n^2
\\
(\delta(\alpha,b), 1) & i = n^2
\end{cases}
\]
It is not hard to verify that $A'_2$ satisfies (iii).
\proofbox

\medskip

\noindent In the next corollary, a {\em Junta} of size $t$ is a function $h:\{-1, 1\}^n\to \{0,1\}$ that depends on $\le \log(t)$ variables. The rational behind this definition of size, is the fact that in order to describe a general function that depends on $K$ variables, at least $2^K$ bits are required. Likewise, the sample complexity of learning Juntas is proportional to $t$ rather than $\log(t)$

\begin{corollary}\label{cor:juntas}
For every constant $q_0>0$, $q_0$-local queries cannot help to learn poly-sized Juntas.
\end{corollary}
\proof
Consider the mapping $\varphi:\{-1, 1\}^n\to \{-1, 1\}^{(2q_0+1)n}$ that replicates each coordinate $2q_0+1$ times. To establish the corollary, we show that $\varphi$ is a $q_0$-reduction of type B from poly-sized Juntas to poly-sized Juntas. Indeed, let $h:\{-1, 1\}^n\to \{0,1\}$ be a function that depends on $K$ variables. It is enough to show that there is a function $h'$ on the variables $\{z_{i,j}\}_{i\in [n], j\in[2q_0+1]}$ that satisfies (i) $h=h'\circ\varphi$, (ii) for every $\x\in\cx$, if $\z\in \{-1, 1\}^{(2q_0+1)n}$ is obtained from $\varphi(\x)$ by modifying $\le q_0$ coordinates, then $h'(\z) = h'(\varphi(\x))$ and (iii) $h'$ depends on $(2q_0+1)K$ variables. 
It is not hard to check that the following function satisfies these requirements:
\[
h'(\z) = h\left(\maj(z_{1,1},\ldots,z_{1,2q_0+1}),\ldots,\maj(z_{n,1},\ldots,z_{n,2q_0+1})\right)
\]
\proofbox

\medskip

\noindent We remark that by taking $q_0=\log^{1-\epsilon_0}(n)$ for some $\epsilon_0>0$, and using a similar argument, it can be shown that an efficient algorithm for learning poly-sized Juntas with $\left(\log^{1-\epsilon_0}(n)\right)$-local queries would imply a PAC algorithm for poly-sized Juntas that runs in time $n^{O\left(\log^{1-\epsilon_0}(n)\right)}$.

\begin{corollary}\label{cor:trees}
For every constant $q_0>0$, $q_0$-local queries cannot help to learn poly-sized decision tress.
\end{corollary}
\proof
As with Juntas, consider the mapping $\varphi:\{-1, 1\}^n\to \{-1, 1\}^{(2q_0+1)n}$ that replicates each coordinate $2q_0+1$ times. To establish the corollary, we show that $\varphi$ is a $q_0$-reduction of type B from poly-sized decision trees to poly-sized decision trees. Indeed, let $h:\{-1, 1\}^n\to \{0,1\}$ be a function that is realized by a decision tree $T$. It is enough to show that there is a function $h'$ on the variables $\{z_{i,j}\}_{i\in [n], j\in[2q_0+1]}$ that satisfies (i) $h=h'\circ\varphi$, (ii) for every $\x\in\cx$, if $\z\in \{-1, 1\}^{(2q_0+1)n}$ is obtained from $\varphi(\x)$ by modifying $\le q_0$ coordinates, then $h'(\z) = h'(\varphi(\x))$ and (iii) $h'$ can be realized by a tree of size $|T|^{2q_0+1}$.
As we explain, this will hold for the following function:
\[
h'(\z) = \maj\left(h(z_{1,1},\ldots,z_{n,1}),\ldots,h(z_{1,2q_0+1},\ldots,z_{n,2q_0+1})\right)
\]
It is not hard to check that $h'$ satisfies (i) and (ii). As for (iii), consider the following tree $T'$. First replicate $T$ on the variables $z_{1,1},\ldots,z_{n,1}$. Then, on the obtained tree, replace each leaf by a replica of $T$ on the variables $z_{1,2},\ldots,z_{n,2}$. Then, again, replace each leaf by a replica of $T$ on the variables $z_{1,3},\ldots,z_{n,3}$. Continues doing so, until the leafs are replaced by a replica on the variables $z_{1,2q_0+1},\ldots,z_{n,2q_0+1}$. Now, each leaf in the resulted tree corresponds to $(2q_0+1)$ root-to-leaf paths in the original tree $T$. The label of such leaf will be the majority of the labels of these  paths.
\proofbox

\medskip

\noindent As with Juntas, we remark that by taking $q_0=\log^{1-\epsilon_0}(n)$ for some $\epsilon_0>0$, and using a similar argument, it can be shown that an efficient algorithm for learning poly-sized trees with $\left(\log^{1-\epsilon_0}(n)\right)$-local queries would imply a PAC algorithm for poly-sized trees that runs in time $n^{O\left(\log^{1-\epsilon_0}(n)\right)}$. 

In the sequel, we say that a function $h:\{-1, 1\}^n\to\{0,1\}$ is realized by a poly-sized polynomial, if it is the function induced by a polynomial with polynomially many non-zero coefficient and degree $O\left(\log(n)\right)$. A similar convention applies to the term poly-sized polynomial threshold function. We remark that the following corollary and its proof remain correct also if in our definition, we replace the number of coefficients with the $\ell^1$ norm of the coefficients vector.
\begin{corollary}\label{cor:sparse_pol}
For every constant $q_0>0$, $q_0$-local queries cannot help to learn poly-sized polynomials, as well as poly-sized polynomial threshold functions.
\end{corollary}
\proof
We will prove the corollary for polynomials. The proof for polynomial threshold functions is similar.
Consider the mapping $\varphi:\{-1, 1\}^n\to \{-1, 1\}^{(2q_0+1)n}$ that replicates each coordinate $2q_0+1$ times. To establish the corollary, we show that $\varphi$ is a $q_0$-reduction of type B from poly-sized polynomials to poly-sized polynomials. Indeed, let $h:\{-1, 1\}^n\to \{0,1\}$ be a poly-sized polynomial. It is enough to show that there is a poly-sized polynomial $h'$ on the variables $\{z_{i,j}\}_{i\in [n], j\in[2q_0+1]}$ that satisfies (i) $h=h'\circ\varphi$, and (ii) for every $\x\in\cx$, if $\z\in \{-1, 1\}^{(2q_0+1)n}$ is obtained from $\varphi(\x)$ by modifying $\le q_0$ coordinates, then $h'(\z) = h'(\varphi(\x))$.
As we explain next, this will hold for the following polynomial:
\[
h'(\z) = h\left(\maj(z_{1,1},\ldots,z_{1,2q_0+1}),\ldots,\maj(z_{n,1},\ldots,z_{n,2q_0+1})\right)
\]
It is not hard to check that $h'$ satisfies (i) and (ii). It remains to show that $h'$ is poly-sized. Indeed, the majority function on $2q_0+1$ coordinates is a polynomial of degree $\le 2q_0+1$ with at most $2^{2q_0+1}$ non zero coefficients (this is true for any function on $2q_0+1$ coordinates). Hence, if we replace each variable in $h$ by a polynomial of the form $\maj(z_{i,1},\ldots,z_{i,2q_0+1})$, the degree is multiplied by at most $(2q_0+1)$, while the number of non zero coefficients is multiplied by at most $2^{2q_0+1}$.
\proofbox

\medskip

\noindent As with Juntas and decision trees, by taking $q_0=\log^{1-\epsilon_0}(n)$ for some $\epsilon_0>0$, and using a similar argument, it can be shown that an efficient algorithm for learning poly-sized polynomials or polynomial threshold functions with $\left(\log^{1-\epsilon_0}(n)\right)$-local queries would imply a PAC algorithm for the same problem that runs in time $n^{O\left(\log^{1-\epsilon_0}(n)\right)}$.

\section{Conclusion and Future Work}
We have shown that in the distribution free setting, for many hypothesis classes, local queries are not useful. As our proofs show, this stems from the fact that learning these classes without queries can be reduced to a case where local queries are pointless, in the sense that the answer to them is either always $1$, or the label of the closest training example.
On the other hand, the learning problem of DNFs with evident examples circumvents this property. Indeed, the underlying assumption enforces that local changes can change the label in a non-trivial manner. While this assumption might be intuitive in some cases, it is certainly very restrictive. Therefore, a natural future research direction is to seek less restrictive assumptions, that still posses this property.

More concrete direction arising from our work concern classes for which we have shown that $\left(\log^{0.99}(n)\right)$-local queries are unlikely to lead to efficient algorithms. We conjecture that for some $a>0$ even $\left(n^a\right)$-local queries won't lead to efficient distribution free algorithms for these classes.

\bibliography{my_bib}

\end{document}